# Applying Chatbots to the Internet of Things: Opportunities and Architectural Elements

Rohan Kar[1]
Hyderabad, India
rohankar99@gmail.com

Rishin Haldar[2]
School of Computing Sciences and Engineering,
VIT University,
Vellore, India
rishinhaldar@vit.ac.in

*Abstract*—Internet of Things (IoT) is emerging as a significant technology in shaping the future by connecting physical devices or things with internet. It also presents various opportunities for intersection of other technological trends which can allow it to become even more intelligent and efficient. In this paper we focus our attention on the integration of Intelligent Conversational Software Agents or Chatbots with IoT. Literature surveys have looked into various applications, features, underlying technologies and known challenges of IoT. On the other hand, Chatbots are being adopted in greater numbers due to major strides in development of platforms and frameworks. The novelty of this paper lies in the specific integration of Chatbots in the IoT scenario. We analyzed the shortcomings of existing IoT systems and put forward ways to tackle them by incorporating chatbots. A general architecture is proposed for implementing such a system, as well as platforms and frameworks – both commercial and open source – which allow for implementation of such systems. Identification of the newer challenges and possible future directions with this new integration, have also been addressed.

I. INTRODUCTION

The Internet of Things (IoT) is not just a well-recognized phenomenon but one that is shaping the digital age. It introduces an era of interconnected smart objects or 'things' developed upon existing internet architectures. By using unique addressing schemes and standard communication protocols, IoT interconnects these things or objects thereby creating a varied range of technologies are able to interact with each other and reach common goals [1].

An essential goal of connecting various sensors, actuators and services and collecting/processing data from them is to generate situational awareness and enable machines and human users to make sense of themselves and their surrounding environments.

The proliferation of IoT can be seen through adoption of these "smart devices" in our daily life which include applications in Manufacturing, Agriculture, Medical and Healthcare, Transportation, Building and Home Automation and Energy Management among others. A report by Gartner estimates that there will be over 20 Billion connected things in activity by 2020 with Cisco estimating the number to be over 50 Billion [2, 3]. Among them more than half of all IoT endpoints in the consumer space alone. Hence IoT is a phenomenon which is certain to play a major role in our daily interaction with the digitally connected world.

*A. Scope of Internet of Things*

The literature presents various ways to define the Internet of Things. The RFID group defines Internet of Things as "world-wide network of interconnected objects uniquely addressable, based on standard communication protocols". ITU [4] defines it as "a global infrastructure for the information society, enabling advanced services by interconnecting (physical and virtual) things based on existing and evolving interoperable information and communication technologies".

While considering the broad vision of IoT, this paper focuses on the perspective of connected things and applications for those things. To do this we simply create a separation of concern between the fragmented lower Open System Interconnection (OSI) layers of IoT and the unifying adopted upper layers of IoT communication which uses the World Wide Web and its standard network protocols.

The entire IoT system consists of Sensors (temperature, light, motion, etc.), Actuators (displays, sound, motors, etc.), Computation (programs and logic), and Communication interfaces (wired or wireless). However, based on established advantages presented in the literature [5, 6, 7, 8], our scope will be limited to interaction with IoT through Web Application Programming Interfaces (API) and in particular Hypertext Transfer Protocol (HTTP) based Representational State Transfer (REST) Architectures. A popular approach of Web of Things has been illustrated in Fig 1 based on [6].

The Evans Data Corporation (EDC) Report: Internet of Things - Vertical Research Service study [9] reveals that more than half of IoT developers connect to devices primarily through the cloud. The massive growth and acceptance of these cloud based platforms such as IBM IoT Platform, AWS IoT, Microsoft Azure IoT and Cisco IoT show that the new generation of IoT applications concentrate on cloud based platforms with the lower layers (Transfer, Transport and Network). Hence, this paper also proposes the use of IoT cloud based platforms in our architectural design. This is discussed further in Section 4.

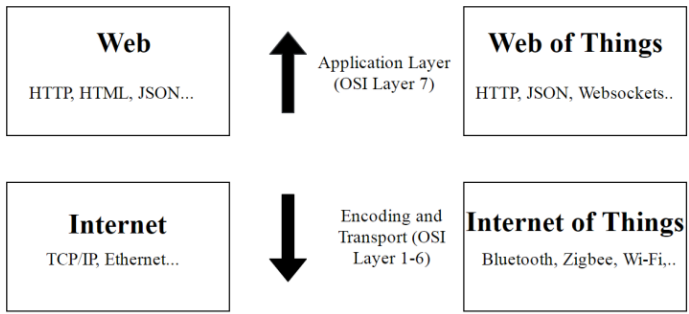

Fig. 1. Web of Things as shown in [6]

*B. Scope of Chatbots*

This paper proposes the use of Intelligent Conversational Agents. We refer to these as simply Chatbots (also known as Chatterbots or bots in general). Interestingly, there are many definitions for Chatbots in close relation with Software Agents (SA), Virtual Agents (VA) or Intelligent Personal Assistants (IPA) in literature and these have often been used in conjunction with each other. The term "Agents" itself has many definitions but among the earliest and most well-known uses of the term is [10] - "A self-contained, interactive and concurrently-executing object, possessing internal state and communication capability."

The scope of Software Agents can be most closely associated with Chatbots and has been well documented in literature [11]. The following key properties have been associated with Software Agents [12]: (1) reactive, (2) pro-active and goal-oriented, (3) deliberative (4) continual (5) adaptive (6) communicative, and (7) mobile. The purpose of this paper is not to explore the various types of Software Agents and agent based systems or its properties but rather propose the solution to challenges faced in IoT through the use of the umbrella term for these Intelligent Conversational Agents, Software Agents or Chatbots as we refer to them. It is also important to note that Software Agent distinguishes itself from Intelligent Agents (also known as Rational Agents). Intelligent agents are not only computer programs. They can also be machines, humans or anything that is capable of a goal directed behavior [13].

Typically Chatbots are classified into two types: (1) Chatbots that function based on Rules (2) Chatbots that function based on Artificial Intelligence. Chatbots that function on rules are often limited as they are only as smart as they are programmed. On the other hand, AI based Chatbots give the impression of being "intelligent" as they are capable of understanding natural language, not just pre-defined commands but get smarter as they interact more due to their ability to maintain different states. Based on this, concepts such as Virtual Agents and Intelligent Personal Assistants (IPA) have come up, which uses natural language processing, as well as speech recognition techniques. For example, Apple Siri, Amazon Alexa, Microsoft Cortana and Google Assistant.

In this paper we present a novel paradigm combining these two disparate concepts in a single solution. However, the studies of these paradigms have largely been separate endeavors. We discuss how using chatbots as intelligent conversational interfaces can be used to address critical problems in IoT. We also propose a high level conceptual architecture and discuss key elements involved in communicating with an IoT system through Chatbots. To explain in the context of real world applicability, we put forth existing solutions to each of the components in the architecture including frameworks, platforms and specify open-source tools which can be used to build such a system.

The rest of this paper is organized as follows: In section II we discuss our motivation for introducing this novel concept of Chatbots in Internet of Things and discuss other literature work that has helped shape this concept. In Section III, we evaluate and examine the shortcomings and challenges of current IoT systems and the opportunity for chatbots to address them. Section IV proposes a system design and the key architectural elements. Finally, we present our concluding remarks by assessing opportunities and scope for future research and development in Section V.

## II. MOTIVATION AND BACKGROUND STUDY

The key to the massive adoption and diffusion of IoT is the proliferation of Internet in our daily lives. We use the internet to search for information, check emails, consume media, and connect with people via social networks and so much more. With around 40% of the global population (3.4 Billion) currently using the world wide web, this number is estimated to increase to 7.6 billion global internet users in 2020, a majority of which use mobile devices (phones, tablets, wearables etc.) [14]. Hence the internet has played a vital role as a global backbone for information sharing, interconnection of physical objects with computing/networking capabilities for applications and services spanning numerous use cases. Internet alone, however, cannot address all issues of IoT. First, we will briefly discuss the challenges in IoT, and then mention the motivation for choosing intelligent conversational interfaces.

*A. Challenges in IoT*

Despite the wide scale efforts to popularize IoT, it still offers many practical challenges. Primarily, IoT systems operate in isolated technology or vendor specific silos which inhibits capability, value, and interoperability and create a widely disparate area [15]. Specifically, by restricting heterogeneous devices (home appliances, mobile phones, embedded devices etc.), sensors and services to communicate with each other across interconnected networks, possibilities of countless applications are hindered.

Secondly, the sheer number of connected things has already started to create problems in application, device and data management in IoT [16]. To address this issue, IoT platforms (such as Cisco IoT, IBM IoT, Microsoft Azure IoT, AWS IoT) offer scalable, distributed cloud based services in order to allow businesses to quickly connect to an established infrastructure, service or software without having to worry about the backend complexities. While IoT Cloud is a step in the right direction, offering many advantages, it still presents many challenges particularly in interoperability which has led to the issues of platform fragmentation [17, 18].

IoT systems also face a challenge of unifying User Interfaces (UI). It becomes increasingly difficult on users to keep track and access multiple applications, dashboards for every new "IoT object" in their ecosystem [19]. Hence unifying experiences across multiple connected things and providing them with a high degree of smartness for improved user experience is a key challenge.

*B. Relevance of Conversational User Interfaces*

According to reports, Chat interfaces which are used in Instant Messaging (IM) platforms (Such as Facebook Messenger, Slack, Kik, and Telegram) have been immensely popular and continue to show steady growth. IM services have more active users than any other internet application including social networks, mailing applications [20]. This report shows how the top ten messaging platforms alone account for nearly 4 Billion users. The global acceptance of chat based interfaces allows for ease of adoption and diffusion of newer technologies built on top of the pre-existing platforms (such as Chatbot Applications). Therefore the global proliferation of chat platforms only furthers the motivation to develop interesting applications and use cases with chatbots.

On a different note, advancements made in the areas of Artificial Intelligence, especially Natural Language Processing have furthered the efficiency and quality of Chatbots in terms of its conversational simplicity, easy adaptability and capabilities in allowing the user to make complex requests through simple natural language.

However, the similarity of both Chatbots and IoT lies in adoption of their services through relatively simple, often RESTful Web APIs. The rationale behind it has been based on the following observations:
(1) Developers can take an API or service-oriented approach to development for both IoT as well as Chatbots. This means that application development methodologies would be the same with both embedded devices as with any web service (including Chatbots) that use Web APIs and in particular using RESTful architectures.
(2) Chatbot applications just like IoT applications can be designed and deployed on cloud platforms which allow simple development and deployment without concern about the underlying technologies such as Transport layer, Storage, Processing etc.
(3) Owing to HTTP RESTful standards and protocols, it becomes technologically feasible and simple to integrate chatbot applications into IoT systems using application layer as the only concerned medium

This ease of integration is a key motivation to develop platforms and frameworks which can synchronize chatbot applications within IoT platforms and frameworks.

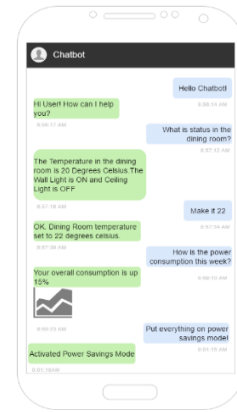

Fig. 2. Sample of a IoT Chatbot-User conversation

### III. OPPORTUNITIES FOR CHATBOTS IN IOT

The shortcomings of modern IoT systems can be broadly classified into two types: (i) Technology Centric Challenges and (ii) Human Centric Challenges. With the help of sample chatbot-user conversations given in Fig. 2, we discuss the opportunities for Chatbots and demonstrate ways in which Chatbots can overcome challenges in IoT.

**Use Case (A)**
User: "Keep the living room temperature comfortable"
**Chatbot:** "The weather outside is a cool 17 degrees Celsius. Setting temperature in the living room to 21.4 degree Celsius."
**Use Case (B)**
User: "How much is my car charged"
**Chatbot:** "The Tesla Model S is currently 40% charged. 3 Hours 10 minutes to full charge."
**Use Case (C)**
User: "Turn the light on in the guest bedroom"
**Chatbot:** "Which light would you like to have turned ON? The Lamp or Table Light?"
User: "Both"
**Use Case (D)**
User: 'Help me setup my new device'
**Chatbot:** "Here is some help to guide you through the setup"
"Which device would you like to setup?
 1) Smart Lock 2) Smart Kettle 3) Smart light?"
User: 1
**Chatbot:** "Ok, Enter your secret passcode for the smart lock"
User: "*****"
**Chatbot:** "Done. Smart Lock is now setup."
**Use Case (E)**
**Chatbot:** "The monitoring service indicates that the smart lock has been offline for over 24 hours."
**Chatbot:** "Would you like me to report the issue to the Smart Lock Vendor?"
User: "No, I want to talk to a human"
**Human-Operator**: "I can see the issue you are facing. I will try to resolve it remotely."

*A. Technology Centric Challenges of IoT*

   *1) Data Management*

   A key challenge in the realm of IoT, is managing the vast amount of big data being generated, as IoT sensors are becoming easily affordable. Not only is the data generated by the sensors large but also diverse (varying in quality and type) and multimodal (temperature, light, sound, video, etc.) in nature. While data deluge is one challenge, drawing insights from the data and being able to present it in a timely, understandable way is a much larger challenge.

The situation can be best illustrated by the well-known Knowledge Hierarchy also called the DIKW (Data, Information, Knowledge, Wisdom) Pyramid in the context of IoT [21, 22] which calls for solutions to tackle the difficult challenges as one moves up in the pyramid. The data gets smaller but becomes more difficult to gain abstractions and perceptions (Knowledge), which is required to derive actionable intelligence (Wisdom). Chatbots are attempting to solve the problems of data and information management by mainly addressing the upper layers of the DIKW pyramid.

#### a) Data Context

Processing and analyzing of IoT data can be solved through the many "big data" solutions and cloud platforms which offer storage and computing infrastructure to accomplish the task. These existing IoT cloud solutions handle Data source and transmission challenges. However, a major challenge of existing IoT systems is conveying data about the various interconnected devices (sensors, objects etc.) back to the user in a simple humanly understandable way. This requires context, which is achieved by enabling Chatbots to understand the true intent of the user query and collect and process information from their environments. Moreover, Chatbots have access to a global network of information via the internet and can be easily programmed to retrieve information in real-time which can improve the context.

In practical terms, Chatbots simplify the way we consume information from multiple screens and heavy data and graphics to simple conversational interfaces capable of delivering highly contextual and intelligible information within the flow of the chat app itself. Achieving this high-level of abstraction can deliver actionable intelligence (wisdom) with domain and user knowledge to maximize the full potential of IoT. For example, in use case (A), the query was relatively vague. The Chatbot could have used contextual information from Real time outside temperature along with knowledge from historical user preferences to perform a specific action.

#### b) Information Retrieval

IoT dashboards are often saturated with various metrics, data points, charts and tables making it difficult for users to find the required information. Chatbots can effectively solve this problem by responding quickly to direct queries with highly accurate information. By understanding the specific intent of the user they limit the scope of information for presentation. In terms of the knowledge hierarchy, Chatbots perform lookup and abstraction on IoT data. By automatically providing IoT data as well as user-contextual data to an analytics, Chatbots can also derive its own knowledge. For example, in use case (B) the query only asked for Battery Charge related information and nothing else. The Chatbot limited the response accordingly.

### 2) Device and Application Management

A key challenge of IoT has been the fragmentation of technology [17, 18]. Having application interoperability between heterogeneous devices from a single remote (mobile device or operation terminal) is especially uncommon. For Example, consider the situation where a smart light and a Heating Ventilation and Air Conditioning (HVAC) system belong to the same network and environment yet may have different user control terminals which are mutually independent entities, unaware of each other nor able to control or communicate with each other.

Chatbots are built on IM platforms (such as Facebook Messenger and Slack) which support multiple different chatbot applications. A single chatbot application as well can use unique HTTP REST APIs pertaining to different IoT devices. Chatbots can thus act as a single interface for communication between single purpose devices (eg. Controlling two smart lights), heterogeneous devices (eg. Controlling a HVAC and a Smart Car) and even different IoT ecosystems (eg. Controlling Smart home devices and Smart Retail devices) in the case of cloud based IoT. For example, in the Use cases above, the same chatbot is being used to converse with multiple heterogeneous devices. Provided the right permissions are available it can communicate with Public IoT devices.

### 3) Bridging Data across Platforms and Services

IoT platforms can be seen as software development environments which handle Device management, Application management, Connection Management, Dashboard and Analytics. Yet owing to platform fragmentation [17, 18], sharing of data across platforms is still uncommon. One solution is to solve the issue at the application level by using 3rd party services, which through APIs can access data from each platform. The data can be either processed on the various platforms or extracted into another service and used to deliver something of value which can then be presented through a single Chatbot interface.

### 4) Search and Discoverability

A key attribute in IoT is the natural tendency of objects to be dispersed in the environment while being interconnected and identifiable at class-level (i.e. common information across the same class) or serial-level (i.e. unique to an individual object) [23].

Based on the permissions of the requester and the availability of the connected objects in the scope of the environment, IoT requires lookup and discovery services to effectively find and control these objects. Such services include availability of sensors and actuators which the Chatbot would be able to retrieve from the entities and convey to the user at the appropriate times.

*5) Monitoring and Reporting*

From IoT wearables such as health monitoring devices to industrial sensors which convey information in real time, monitoring and reporting are key aspects of IoT systems. Chatbots can also be effectively used to perform as monitoring services by integrating with solutions such as Application Performance Management (APM). Accessing data from various IoT systems is a key advantage which is unique to Chatbots in this scenario. Similarly Chatbot services can utilize its own reporting services and present the abstracted information to the user in an actionable and timely manner.

*B. Human Centric Challenges of IoT*

Chatbots were created with the primary purpose of improving the human-computer user experience. As such, solving the user experience shortcomings of IoT systems can be an important opportunity for chatbots. IoT, with its complex system of applications, sensors, actuators and services presents a daunting challenge of gaining technical knowledge to interact with these various components. Hence exposing settings and configurations to users presents an obvious and unfriendly burden that is far from ideal.

*1) Cognitive Burden*

The changing technology landscape of IoT is both imminent and rapid. Furthermore, as newer features and use cases are introduced, there is an added responsibility to educate the end users which can be burdensome for both the users and the developers of the system. Complicated systems cause difficulties in adoption and diffusion. As an assistive technology, chatbots can simplify the learning curve by the following ways:

*a) Help Features:* IoT enabled Chatbots can feature help texts which clarify the user request to ensure that the action performed is same as the one intended.

*b) Recommendations:* Chatbots can recommend possible actions to the user which can be made more intelligent and context aware depending on user preferences and the dynamics of the environment.

*c) Automation:* Chatbots are good at automating common cyclic, tasks and can perform certain actions such as monitoring availability of sensors (uptime, downtime etc.) and others through routine API calls, Websockets or Publisher-Subscriber methods.

*d) Better Quality of Service (QoS):* Feedback loops can be easily integrated within chatbots to aggregate most frequent queries and data from the process can be used to improve the future Quality of Service (QoS).

As more use cases are discovered, chatbots can make the adoption and diffusion of IoT systems significantly easier and reduce the cognitive burden required to understand the functionalities of these systems.

*2) User Interface Opportunities*

Graphical User Interfaces (GUI) for IoT are largely functional in nature. While it achieves simplicity by displaying virtual switches, sliders and buttons rather than passing complex commands, it still has some shortcomings which Chat interfaces can solve. (1) Chat interfaces understand natural language which makes interaction with the system as simple as asking queries and receiving answers. There is no need for navigation of menus and finding the right icon/button to perform a task. (2) Chatbots use machine learning techniques to learn about an individual user and can personalize the service to that user. In this way, they can understand the unique way the user converses with while maintaining the natural flow of the conversation (3) They are also highly contextual interfaces and can understand the intent in the scope of the past interactions which is unique to chat based interfaces (also speech). (4) Chat based interfaces concern mostly textual information thereby simple log files can be maintained and consequently analyzed to make debugging easier.

*3) Configuration Challenges*

Apart from the knowledge required to adapt to the new systems and ease cognitive burden, each IoT device has its own unique setup and configuration in terms of software, network, firmware etc. As the number of different IoT devices increase, it becomes difficult and burdensome at best to navigate the interfaces of various applications and properly configure the system. Often technicians are involved to configure and explain the uses of the system.

Using Chatbots, users can be guided and advised on the right configurations for their system by creating step-by-step setup processes using predefined configuration APIs. This also reduces human effort in setting up the system. For example: A new device was configured in use case (D).

*4) Lack of Automated Error Reporting*

The distributed nature of most IoT systems implies that user report databases of IoT errors are spread across multiple organizations, Operating System (OS) vendors, ISPs, and device vendors which makes automated problem reporting a major challenge. Furthermore, users themselves are uncertain which organization to report the particular issue to. Thus, various stakeholders in the system have a limited understanding of the true nature of the problem and avoid sharing information with each other. Chatbots, in this scenario, can access these reported problems and by integrating other services, be able to not only retrieve information from the IoT system but send information to it. In Use Case (E), the chatbot identified the correct stakeholder to send the error.

*5) Support Challenges*

Remedying hardware and software issues in modern consumer IoT systems can be an irksome task. The recourse is to call the service provider for technical support or in many cases return the product. Either way it is an unnecessary burden on the user as well as the support vendors in today's cost structure.

Smart Chatbots often have support services built into their functionality. It can even integrate human-in-the-loop systems to handle situations the Chatbot is not trained or authorized to perform, in real-time. In this manner, users need not go beyond the scope of the chatbot application to look for product support. Any software issue or hardware malfunction can be monitored and Over the Air (OTA) software repairs can be performed. Chatbots can also be used to schedule technical repairs making it a convenient and fast solution to customer support [24]. For example, in Use case (E), a human operator was made to intervene.

## IV. SYSTEM DESIGN AND ARCHITECTURAL ELEMENTS

We present a conceptual system design which will aid in building Chatbot systems for IoT. Fig. 3 presents the high level view of the overall architecture consisting of the IoT system and the Chatbot system.

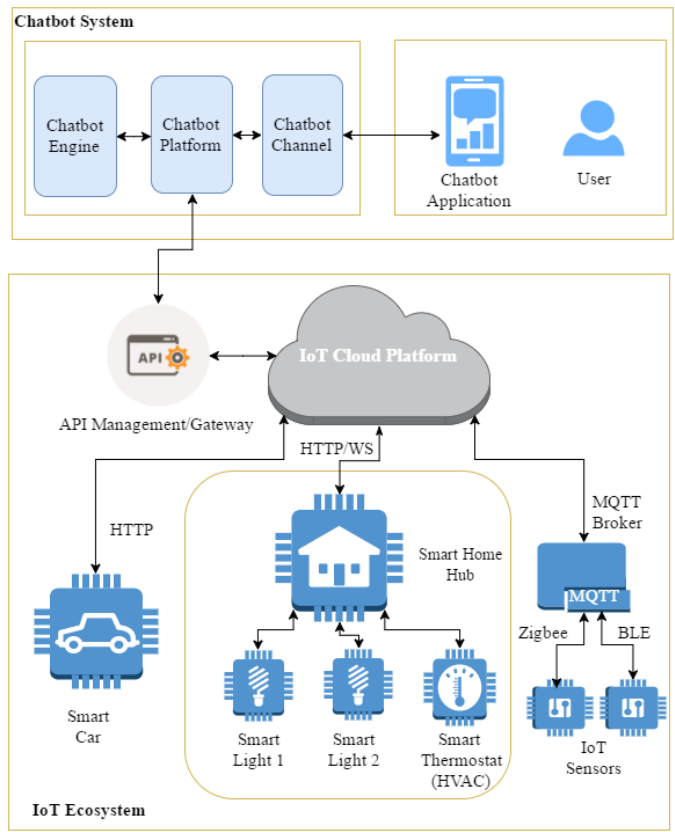

Fig. 3. Proposed System Design of IoT-Chatbot System

### A. IoT System

*1) IoT Devices*

In the context of the paper, we refer to an IoT device in the broad scope of the term as a "uniquely identifiable IoT endpoint which can be accessed and controlled using RESTful Web APIs". In the situation an embedded device does not have APIs, there are existing solutions to easily create APIs for them. For Example: Using platforms such as Zetta, one can create IoT cloud based systems with full-fledged APIs. In the presented system we consider a Home automation system consisting of Smart lights (eg. Philips Hue) and Smart HVAC (consisting of a smart thermostat eg. Nest) as well as a Connected Car or Smart Car (eg. Tesla). However, on principle, any IoT device may be considered for interfacing with Chatbots.

*2) IoT Cloud Platform*

IoT Cloud based Platforms is an important enabling technology in many IoT systems today. They deal with various fragmented technologies in embedded devices from access protocols (eg. Message Queuing Telemetry Transport (MQTT), HTTP etc.), wireless protocols (Zigbee, Bluetooth Low Energy (BLE)) to various services, SDKs and integrations. There has been positive reports on the established advantages of Cloud based IoT platforms [25]. Our system design stresses on accessing and controlling the embedded devices in question, such as Smart Car, Light, Thermostat etc., through the API Management/Gateway of the IoT Cloud Platform, regardless of the standards and protocols of the individual embedded devices. On the other hand, complex components in Networking and Computing infrastructure for the Cloud platform have not been addressed in this paper. Popular IoT platforms today include Microsoft Azure IoT, IBM IoT, APIGEE IoT, Cisco IoT, among others.

### B. Chatbot System

*1) Chatbot Channels and Platforms*

Chatbot Channels are applications which run Chatbots on supported Mobile devices (eg. Smartphones, Tablets) or Terminals (eg. Desktop Applications).They are typically built on top of the existing instant messaging platforms. Popular Chatbot channels include Facebook Messenger, Slack, Telegram, Kik, Skype, Line and Twilio SMS. These channels are essentially the Chatbot applications in which a user interacts with the bot. The area of Chatbot development is still in its infancy and there can be many different architectural approaches in implementing Chatbots.

In some approaches, the channels are interfaced separately from the Chatbot Platforms through connectors. Here, the Chatbot Platforms are hosted on cloud services which can use Webhooks to communicate with the Channel. In the context of this paper, we consider the integration of Chatbots as text based inputs to IoT. By using Software Development Kits (SDK) it is possible to integrate IoT to voice/speech based Intelligent Personal Assistants such as Amazon Echo (using Alexa SDK) and Google Home (using Google Assistant SDK).

## 2) Chatbot Engine

Perhaps the most important component of a Chatbot is the engine, often referred to as Natural Language Understanding (NLU) engine. It is responsible for translating natural language into machine understandable action. Chatbot engines are often highly complex, using various Natural Language Processing (NLP) models and Machine Learning (ML) techniques to provide acceptable levels of accuracy. To make it easier for Chatbot developers, many companies offer the processing capability of the Chatbot engine as a Software-as-a-Service(SaaS) or 'AI-as-a-service' which are applied to Chatbot applications using APIs. For example, Wit.ai and Microsoft LUIS.

This paper is primarily focused on listing the relevant key functional components of the engine in the context of IoT, not on designing the NLP techniques for the Chatbot engine. Next, we include key concepts typically associated with chatbot engines [26, 27]:

### a) Entities:

Entities are domain specific information extracted from the utterance that maps the natural language phrases to their canonical phrases in order to understand the intent. They help in identifying the parameters which are required to take a specific action. To train the chatbot engine, entities which are expected to give the same actions are typically grouped together. Common entities can be predefined as they can be used in many different scenarios. For example: Money, Color, Date time, Location, Number. Domain specific entities can be trained to recognize similar phrases. IoT devices are one such domain specific entity.

For example, for the utterance: "Thermostat", the acceptable phrases may be trained to recognize "Thermostat", "heat", "heating", "AC", "air conditioning" and will be decoded as {"type":iot, "device": "Thermostat"} where the entity is IoT. Entities may also have its own attributes. For example, the Utterance: "$15" can be decoded as {"type": "money", "amount":15, "currency": "dollars"} where the entity is Money in JSON format.

### b) Context:

Determining the Context of the current user expression is an important feature of modern Chatbots. It can be used to handle situations where the utterances may be vague and have multiple meanings depending upon the history of the conversation. Contexts represent the ability of agents to maintain state (also called lifespan or the number of utterances after which the context will be removed) and match the required intent. They may also use information from external sources. For example, if the user was asking about the living room (location) in the first utterance and then after the passing of a few more utterances, mentions a vague statement such as: "make it colder", the Chatbot uses context to understand the earlier reference to the temperature of the living room.

### c) Intents:

Intents are the crux of conversational UI in chatbots. The intents represent what the users are looking to accomplish: get status updates, turn on/off devices, ask for help etc. The message passed from the user (utterance) in natural language is first analyzed for the intent. This means mapping a phrase to a specific action that should be taken by the IoT system as well as the specific dialog to be returned from the Chatbot. The information contained in an intent would be the context and action.

### d) Action:

Action refers to the steps that the IoT device will take when the intent of the user input is recognized. Actions have specified parameters which categorize details about it. Actions will be triggered once they are recognized by the intent For example the actions in a Smart Home may be smartHome.lightsOn, smartHome.doorLock, and smartHome.getStatus. In this scenario, other parameters may also be defined such as location (eg. Dining room), time start/end (eg. 10am, Thursday etc.), schedule (eg. every hour, every minute) etc.

Once the action has been set and the require parameters have also been defined, the correct intent can be mapped to an IoT API endpoint and a HTTP request is made.

Chatbots can also be built using existing frameworks simplify the end-to-end process of creating and integrating Chatbots into messaging and IoT platforms. The key advantages of using a Chatbot framework are: (1) Ease of Development from pre-defined actions, integrations and SDK support for various IoT systems (2) Ability to 'write once deploy anywhere' through integrations with multiple Chatbot Channels. (3) Ability to use AI-as-a-service. For example, LUIS.ai in the case of Microsoft Bot Framework. Other popular Chatbot frameworks include API.ai, Microsoft Bot Framework and IBM Watson Conversation Service.

## V. FUTURE RESEARCH AREAS AND CONCLUSION

IoT is poised to become intrinsic to the day to day activities in the future, however the dynamic nature of IoT has its share of difficulties, and this paper has put forward the concept of using Chatbots to address some of the challenges in IoT. Through this initial endeavor we can identify possible areas that can be worked on for the future which show great potential:

### 1) Stronger AI based Agents

As more advances are made in the field of AI, Software Agents will also grow to become more intelligent in the future. The goal of Strong AI has been to match the machine's intellectual capability to a human being. Immediate research challenges includes improving decision making ability to create more autonomous Chatbots and better NLP as well as Natural Language Generation models to create more natural flows of conversation between humans and bots. Chatbots will play an important role in the research areas of Intelligent Agents as well as Machine-to-Machine (M2M) research in IoT.

### 2) Cyber Physical Systems and IoT

Cyber Physical Systems refer to more advanced, next generation embedded Information and Communications Technology (ICT) systems. They share many similarities with IoT but with higher combination and coordination between physical and computational elements [28]. The US National Science Foundation (NSF) identified cyber-physical systems among the key research areas in the foreseeable future [29]. IoT will play a major role in the transition to CPS as one of the key enabling technologies [16, 30]. Further advancements in AI aspects of Chatbots in IoT will be closely related to the Conversion, Cyber, Cognition and Configuration levels of the 5C CPS Architecture [31].

### 3) Wisdom of Things

The progress of Chatbots in IoT introduces the paradigm of human-in-the-loop systems which has exciting research challenges in the areas of Wisdom of Crowds. The concept of wisdom of crowd suggests that aggregation of information can result in decisions that are better than what could have been achieved by any individual in the group [32]. In the context of IoT, the sharing of big data from billions of sensors and devices creates more value in the ecosystem as compared to not sharing. However it requires data interoperability rather than simply accumulating multiple disparate data sources which are incompatible or have no similarities. Hence Chatbots in IoT systems can use techniques such as Human Swarming, an approach that uses real-time feedback loops from groups of users to make accurate insights. There are plenty of interesting research opportunities in acquiring accurate values from the crowd.

### 4) Evolution of the Semantic Web

As the Internet itself changes, there are many more opportunities for exciting research in the areas of IoT and Software Agents. The development of a Web 3.0 or Semantic Web and its impact on the future of Software Agents has been clearly described in the literature [33]. This poses a great opportunity for research in IoT in terms of Semantic interoperability which can have major impact on the IoT paradigm itself. The evolution of the Semantic Web will have major impact on areas in pervasive computing, M2M technologies resulting in Software Agents being able to draw more value and achieve a higher level of wisdom than before.

Development in the field of IoT has been phenomenal in recent times. Similarly, Chatbot systems are also becoming more intelligent and sophisticated as the days progress. To the best of our knowledge, no work has been published detailing the specific integration of Chatbots to IoT. This paper has attempted to integrate these two fields together by enlisting the key architectural components required and envision possible ways to address some of the present challenges in IoT. We hope that this endeavor will lead to more intelligent and effective integrated IoT systems.